\pdfoutput=1

\documentclass[11pt]{article}

\usepackage{EMNLP2023}

\usepackage{times}
\usepackage{latexsym}
\usepackage{booktabs}
\usepackage{multirow}
\usepackage{caption}
\usepackage{subcaption}

\usepackage[T1]{fontenc}
\usepackage[utf8]{inputenc}
\usepackage{microtype}
\usepackage{inconsolata}

\usepackage{amssymb}
\usepackage{pifont}
\usepackage{fdsymbol}
\newcommand{\clip}{CLIP}
\newcommand{\clipscore}{CLIPScore}
\newcommand{\clipscores}{CLIPScores}

\newcommand{\rmat}{\rho}
\newcommand{\dintinv}{\textsc{DII}}

\newcommand{\cmark}{\ding{51}}%
\newcommand{\xmark}{\ding{55}}%

\usepackage{titlesec}
\titlespacing*{\paragraph}{0pt}{0.5ex}{1ex}

\title{Updating CLIP to Prefer Descriptions Over Captions}

\author{Amir Zur$^{\clubsuit}$ \\
  \texttt{amirzur@cs.stanford.edu} \\\And
  Elisa Kreiss$^{\varheartsuit}$ \\
  \texttt{ekreiss@ucla.edu} \\ \And
  Karel D'Oosterlinck$^{\spadesuit \vardiamondsuit}$ \\
  \texttt{karel.doosterlinck@ugent.be} \\
  \AND
  Christopher Potts$^{\spadesuit}$ \\
  \texttt{cgpotts@stanford.edu} \\ \And
  Atticus Geiger$^{\clubsuit}$ \\ 
  \texttt{atticusg@gmail.com}\\
  \AND
  $^{\clubsuit}$Pr(Ai)$^{2}$R Group \; $^{\spadesuit}$Stanford Univeristy \; $^{\varheartsuit}$UCLA \; $^{\vardiamondsuit}$Ghent University – imec\\
}

\begin{document}
\maketitle
\begin{abstract}
Although \clipscore\ is a powerful generic metric that captures the similarity between a text and an image, it fails to distinguish between a \emph{caption} that is meant to complement the information in an image and a \emph{description} that is meant to replace an image entirely, e.g., for accessibility. We address this shortcoming by updating the \clip\ model with the Concadia dataset to assign higher scores to descriptions than captions using parameter efficient fine-tuning and a loss objective derived from work on causal interpretability.
This model correlates with the judgements of blind and low-vision people while preserving transfer capabilities and has interpretable structure that sheds light on the caption--description distinction.\footnote{Our code is available at \url{https://github.com/AmirZur/updating-clip-concadia/tree/main}.}
\end{abstract}

\section{Introduction}
The texts that accompany images online are written with a variety of distinct purposes: to add commentary, to identify entities, to enable search, and others. One of the most important purposes is (alt-text) \emph{description} to help make the image non-visually accessible, which is especially important for people who are blind or low-vision (BLV) \citep{bigham2006webinsight,morris2016most, gleason2020twitter}. The ability to automatically evaluate descriptions of images would mark a significant step towards making the Web accessible for everyone.

\begin{figure}[t]
    \centering
    \includegraphics[width=0.28\textwidth]{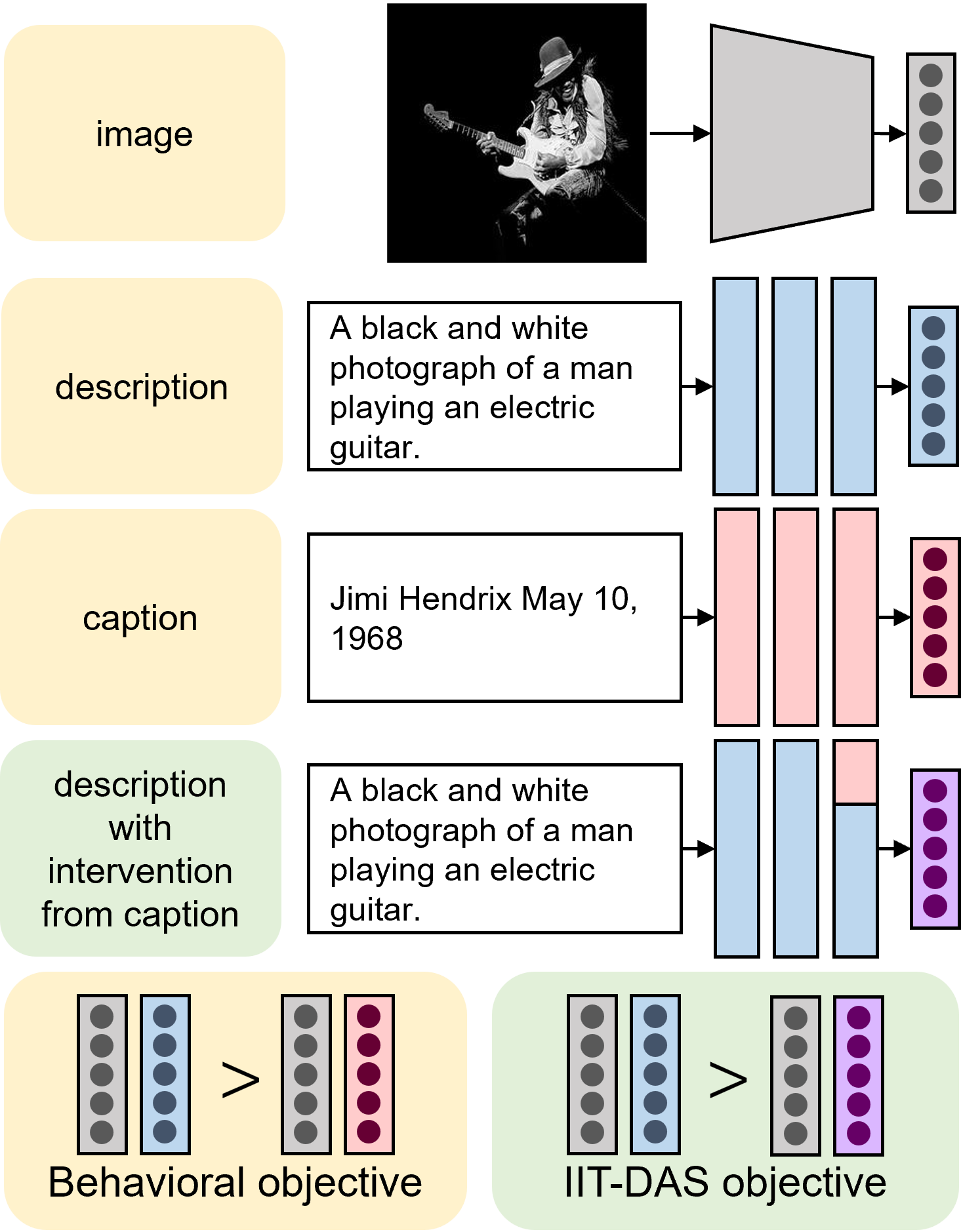}
    \caption{Visualization of a single training step on Concadia, updating \clip\ to prefer descriptions to captions. Though both objectives update \clip\ to be sensitive to the description--caption distinction, IIT-DAS localizes this distinction to a subspace of \clip's activations.} 
    \label{fig:clip-iit}
\end{figure}

Unfortunately, present-day metrics for image-text similarity tend to be insensitive to the text's purpose \citep{kreiss2022concadia}, as they don't distinguish between accessibility descriptions that are intended to replace the image from captions, which supplement them (see Figure~\ref{fig:clip-iit}). Thus current metrics fall short when it comes to making genuine progress towards accessibility \citep{kreiss2022context}. 

The Contrastive Language-Image Pre-training (\clip) model of \citet{Radford} is an important case in point. \clip\ is trained to embed images and texts with the objective of maximizing the similarity of related image--text pairs and minimizing the similarity of unrelated pairs, without reference to the purpose that the text accompanying an image plays. 

The \clipscore\ metric of \citep{Hessel} inherits this limitation; \clipscore\ is \emph{referenceless} in that it makes no use of human-generated, ground-truth texts, but rather depends only on image--text pairs, and so it too is not sensitive to the purpose of the text. Indeed, \citet{kreiss2022context} find that \clipscores\ correlate with neither sighted nor BLV user evaluations of image alt-descriptions. Thus, despite high performance on many image--text classification tasks, \clipscore\ is unsuitable for alt-text evaluation.

The goal of this paper is to update \clip\ to assign higher scores to descriptions than captions when they are both relevant to an image, while preserving the model's ability to select the most relevant text for a particular image. To this end, we fine-tune \clip\ on the Concadia dataset \citep{kreiss2022concadia}, which consists of 96,918 images with corresponding descriptions, captions, and textual context. 

Our core goal in fine-tuning is to teach the model to prefer descriptions over captions. To this end, we use a contrastive loss objective, which updates \clip\ to produce a higher score for the description than the caption of an image in Concadia. In addition, we propose an extension of this objective that seeks to amplify the core distinction and create more interpretable models. The guiding idea here is to  
use Concadia to approximate the counterfactual: \emph{What if the text for this image were a description instead of a caption (or vice versa)?} These counterfactuals enable us to use a novel combination of two recent ideas from causal interpretability research: interchange intervention training (IIT; \citealt{geiger2022inducing}) with a distributed alignment search (DAS; \citealt{geiger2023finding}) to localize the description--caption concept to an activation vector. 

Our experiments lead to a few key findings that have implications beyond the accessibility use case. First, we find that LoRA \cite{hu2021lora} is superior to standard fine-tuning at raising the \clipscore\ assigned to descriptions compared to captions while preserving the original capabilities of \clip. Second, our analysis  shows that improved performance on Concadia results in stronger correlations with BLV user judgements, affirming the value of our update. Third, we find that the IIT-DAS objective results in a more stable fine-tuning process. Fourth, the IIT-DAS objective produces a more interpretable model; to show this, we use \emph{mediated integrated gradients} \citep{sundararajan2017axiomatic,wu2023causal}  to characterize how the description--caption distinction is computed in our fine-tuned models. The key role of the IIT-DAS objective in these results illustrates one way that interpretability research can lead directly to more performant and understandable models.

\begin{table*}
    \centering
    \resizebox{\textwidth}{!}{
    \begin{tabular}{@{} ll| c|ccc|cc @{}}
    \toprule
    & \text{\textbf{Fine-}} & 
    \textbf{Desc > Cap} 
    & \multicolumn{3}{c}{\textbf{Transfer tasks} (F1 Score)}
    & \multicolumn{2}{|c}{\textbf{BLV user eval} (Corr.)} \\
    \textbf{Objective}  &  \textbf{tuning}  & \textbf{Concadia} &  \textbf{Food101} &  \textbf{ImageNet}  &  \textbf{CIFAR100}  &  \textbf{Overall} & \textbf{Imaginability} \\
    \midrule
    None & None & 49.4\% & 76.4\% &  53.6\% &  61.5\% & 0.08 & 0.10 \\
    \midrule
    \multirow{2}{*}{Behavioral}  &  Full  & 90.1\% $\pm$ 0.73 &  27.8\% $\pm$ 24.96  &  14.6\% $\pm$ 13.47  &  30.9\% $\pm$ 24.97  &  0.29 $\pm$ 0.17  &  0.31 $\pm$ 0.19  \\ 
     & LoRA &  90.3\% $\pm$ 0.72 & 64.4\% $\pm$ \phantom{0}8.32  &  42.2\% $\pm$ \phantom{0}6.98 &  \textbf{55.6}\% $\pm$ \phantom{0}2.53  & \textbf{0.36} $\pm$ 0.10  &  \textbf{0.38} $\pm$ 0.12 \\
    \midrule
     \multirow{2}{*}{IIT-DAS} &  Full & 88.9\% $\pm$ 0.80 &  35.1\% $\pm$ 16.35  &  19.5\% $\pm$ \phantom{0}6.77  &  42.8\% $\pm$ \phantom{0}9.85  &  0.24 $\pm$ 0.14  &  0.32 $\pm$ 0.17 \\ 
     & LoRA & 86.6\% $\pm$ 0.84 & \textbf{73.6}\% $\pm$ \phantom{0}3.22  &  \textbf{45.1}\% $\pm$ \phantom{0}3.96  &  53.7\% $\pm$ \phantom{0}3.80  &  0.19 $\pm$ 0.15  &  0.27 $\pm$ 0.15 \\
    \bottomrule
    \end{tabular}
    }
    \caption{The percent of Concadia examples where descriptions are scored higher than captions (column 3), F1 scores on transfer learning tasks (columns 4-6), and correlation between BLV human preferences \cite{kreiss2022context} and model similarity scores (columns 7, 8). The error bounds are 95\% confidence intervals from 5 random seeds. Fine-tuning on Concadia produces models that better correlate with BLV preferences, LoRA is essential for preserving transfer learning, and IIT-DAS sacrifices a modest amount on Desc > Cap for better transfer learning.}
    \label{tab:transfer}
\end{table*}

\section{Related Work}

\paragraph{Image Accessibility}
When images can't be seen, visual descriptions of those images make them accessible. For images online, these descriptions can be provided in the HTML's alt tag, which are then visually displayed if the image cannot be loaded or are read out by a screen reader to, for instance, users who are blind or low-vision (BLV). However, alt descriptions online remain rare \cite{gleason2019s, kreiss2022concadia}.

Image captioning models provide an opportunity to generate such accessibility descriptions at scale, which would promote equal access \cite{gleason2020twitter}. But the resulting models have remained largely unsuccessful in practice \cite{morris2016most,macleod2017understanding,gleason2019s}. \citet{kreiss2022concadia} argue that this is partly due to the general approach of treating all image-based text generation problems as the same underlying task, and instead highlight the need for a distinction between accessibility descriptions and contextualizing captions. Descriptions are needed to replace images, while the purpose of a caption is to provide supplemental information.
\citet{kreiss2022concadia} find that the language used in descriptions and captions categorically differs and that sighted participants tend to learn more from captions but can visualize the image better from descriptions.

\paragraph{Referenceless Text-Image Evaluation Metrics}

We focus on referenceless evaluation metrics \cite{feinglass2021smurf, Hessel,lee2021umic} for text--image models.
These can be applied in diverse contexts and require no human annotations, which makes them valuable tools for rapid system assessment.  
Crucially, such referenceless metrics are context-free. This brings the advantage that they can be applied in a variety of multimodal settings (e.g.~image synthesis, description generation, zero-shot image classification). The downside is that they are generally insensitive to variation in context and purpose. \citet{kreiss2022context,kreiss2023contextref} report progress in incorporating context into these metrics; our work can be seen as complementing those efforts by focusing on textual purpose.

\section{Methods}

Our goal is to fine-tune a \clip\ model $\mathcal{C}^{\theta}$ to prefer a description over a caption when the two texts are relevant to an image, while preserving $\mathcal{C}^{\theta}$'s ability to select the most relevant text for a particular image. To this end, we use Concadia, a dataset that consists of $(x_{\text{image}}, x_{\text{description}}, x_{\text{caption}})$ triplets.
We consider two different contrastive learning objectives and two common fine-tuning methods that minimally update $\mathcal{C}^{\theta}$'s sensitivity to the description--caption distinction. 

\paragraph{Behavioral Objective}
For each triplet $(x_{\text{im}}, x_{\text{des}}, x_{\text{cap}})$, we run $\mathcal{C}^{\theta}$ on the image--caption pair $(x_{\text{im}}, x_{\text{cap}})$ and on the image--description pair $(x_{\text{im}}, x_{\text{des}})$, and update $\mathcal{C}^{\theta}$ to produce a higher score for the latter by minimizing: 
\begin{equation*}
\mathcal{L} =\texttt{CE}\big([\mathcal{C}^{\theta}(x_{\text{im}}, x_{\text{cap}}), \mathcal{C}^{\theta}(x_{\text{im}}, x_{\text{des}})], [0, 1]\big)
\end{equation*}

\paragraph{IIT-DAS Objective}
To better maintain \clip's original capabilities, we 
propose a novel objective called IIT-DAS, which localizes the description--caption distinction in a linear subspace $\mathbf{Z}$ of an activation vector in $\mathcal{C}^{\theta}$. For each triplet, we run $\mathcal{C}^{\theta}$ on the image-caption pair $(x_{\text{im}}, x_{\text{cap}})$, run $\mathcal{C}^{\theta}$ on the image-caption pair again while fixing $\mathbf{Z}$ to the value it takes when $\mathcal{C}^{\theta}$ is run on the image description pair $(x_{\text{im}}, x_{\text{des}})$, and update $\mathcal{C}^{\theta}$ to produce a higher score for the latter by minimizing:
\begin{multline*}
 \mathcal{L}_{\text{IIT}} = \texttt{CE}\big([\mathcal{C}^{\theta}(x_{\text{im}}, x_{\text{cap}}),\\ \dintinv(\mathcal{C}^{\theta}, \rmat^{\theta}, (x_{\text{im}}, x_{\text{cap}}), (x_{\text{im}}, x_{\text{des}}), \mathbf{Z})], [0, 1]\big)
\end{multline*}
where $\rho^{\theta}$ is a randomly initialized orthogonal matrix used to learn the linear subspace $\mathbf{Z}$.
This pushes \clip\ to assign a higher score to a caption with an intervention from a description than to a caption on its own. Likewise, we also train on the objective where descriptions and captions are swapped (prefer a description over a description with an intervention from a caption):
\begin{multline*}
 \mathcal{L}_{\text{IIT}} = \texttt{CE}\big([\mathcal{C}^{\theta}(x_{\text{im}}, x_{\text{des}}),\\ \dintinv(\mathcal{C}^{\theta}, \rmat^{\theta}, (x_{\text{im}}, x_{\text{des}}), (x_{\text{im}}, x_{\text{cap}}), \mathbf{Z})], [1, 0]\big)
\end{multline*}
See Appendix \ref{app:iit} for the formal definition of $\dintinv$.

\paragraph{Fine-tuning} We consider two common fine-tuning methods that update \clip\ in accordance with the selected objective. The first is full fine-tuning where all the parameters of \clip\ are trained with gradient descent.
The second is Low-Rank Adaptation (LoRA), a state-of-the-art fine-tuning technique for transfer learning \cite{hu2021lora}. LoRA training freezes every linear layer $W$ in a model and learns a low-rank matrix $W'$ that is used in unison with the original weights (i.e.~for a residual representation $x$, LoRA computes $Wx + W'x$). 

\section{Experiments}
\subsection{Concadia and Transfer Evaluations}

We use Concadia to quantify the extent to which an image--text model is sensitive to the description--caption distinction.  The metric is simply the proportion of images in Concadia where the description is assigned a higher score than the caption. In the pre-trained \clip\ model, this is the case for $\approx$50\% of the triplets in Concadia (see Table~\ref{tab:transfer}). This means the model assigns higher scores to descriptions only at chance and so is not well suited for the purposes of accessibility. However, we also need to evaluate to what extent each fine-tuning method preserves the original transfer capabilities of \clip. The ideal method will maintain high performance on the image classification transfer tasks, while preferring descriptions over captions. 

We evaluate our fine-tuned models on Concadia and three image classification tasks in disparate domains, selected from \clip's original evaluation suite \cite{Radford} (see Appendix \ref{app:transfer-tasks}).

\paragraph{Results and Discussion}

Table \ref{tab:transfer} shows the performance of each \clip\ model on the Concadia and transfer evaluations. Full fine-tuning is not viable due to its poor performance on the transfer learning tasks. LoRA helps preserve much of the original model's capabilities (dropping $\approx$10\% on Food101 and Imagenet, and $\approx$6\% on CIFAR).

The models trained on the behavioral objective are more sensitive to the description--caption distinction ($\approx$90\%) than IIT-DAS with LoRA ($\approx$87\%). In contrast, models trained on the IIT-DAS objective with LoRA achieve the best performance on transfer tasks (preserving \clip's original Food101 accuracy within its 95\% confidence interval), though sacrificing some sensitivity to the description--caption distinction ($\approx$86\%). The IIT-DAS objective is comparable to the behavioral objective; either one may be preferable depending on the desired balance between description--caption sensitivity and transfer capabilities.

\subsection{BLV and Sighted Human Evaluations}

A preference for descriptions over captions should align \clipscore\ ratings better with the quality judgements of BLV and sighted individuals. 
To evaluate this, we use data from 
\citealt{kreiss2022context}.
\citeauthor{kreiss2022context}\ conducted an experiment where sighted and BLV participants were asked to judge a text describing an image in the context of an article (see Appendix~\ref{app:human-correlations} for experimental details). For our purposes, we ignore the context and isolate the benefit of descriptions compared to captions. Participants rated the quality of the image descriptions along four dimensions, summarized below.

\paragraph{Overall} The overall value of the description as an alt-text description of the image.

\paragraph{Imaginability} How well the participant can visualize the image given the text description. This isn't evaluated for sighted individuals with the image, since they are able to see the reference image.

\paragraph{Relevance} Whether the description includes relevant details from the image, given the context in which the image appears (i.e., the preceding paragraph in a Wikipedia article).

\paragraph{Irrelevance} Whether the description avoids irrelevant details from the image, given the context of in which the image appears.

\begin{table}
\centering
\resizebox{\linewidth}{!}{
\begin{tabular}{l|ll cccc}
\toprule
\textbf{Group}  &  \textbf{Objective}  &  \textbf{Finetuning}  &  \textbf{Overall}  &  \textbf{Imaginability}  &  \textbf{Relevance}  &  \textbf{Irrelevance}  \\
\midrule
\multirow[t]{4}{*}{\emph{BLV}} & None & None & 0.08 & 0.10 & 0.09 & 0.09 \\
\cline{2-7}
&  \multirow[t]{2}{*}{Behavioral}  &  Full	& 0.24{\scriptsize$\pm$ 0.04}	& 0.26{\scriptsize$\pm$ 0.03}	& 0.21{\scriptsize$\pm$ 0.04}	& 0.05{\scriptsize$\pm$ 0.09}  \\ 
&    &  LoRA	& 0.29{\scriptsize$\pm$ 0.03}	& 0.29{\scriptsize$\pm$ 0.01}	& 0.23{\scriptsize$\pm$ 0.02}	& -0.01{\scriptsize$\pm$ 0.03}  \\ 
\cline{2-7}
&  \multirow[t]{2}{*}{IIT-DAS}  &  Full	& 0.29{\scriptsize$\pm$ 0.05}	& 0.34{\scriptsize$\pm$ 0.05}	& 0.30{\scriptsize$\pm$ 0.06}	& 0.07{\scriptsize$\pm$ 0.05}  \\ 
&    &  LoRA	& 0.20{\scriptsize$\pm$ 0.09}	& 0.28{\scriptsize$\pm$ 0.10}	& 0.24{\scriptsize$\pm$ 0.07}	& 0.01{\scriptsize$\pm$ 0.06}  \\
\midrule
\multirow[t]{4}{*}{\emph{Sighted}} & None & None & $-$0.01 & 0.06 & 0.00 & $-$0.17 \\
\cline{2-7}
\emph{(no image)} & \multirow[t]{2}{*}{Behavioral}  &  Full	& 0.22{\scriptsize$\pm$ 0.04}	& 0.13{\scriptsize$\pm$ 0.09}	& 0.17{\scriptsize$\pm$ 0.03}	& -0.03{\scriptsize$\pm$ 0.03}  \\ 
&    &  LoRA	& 0.20{\scriptsize$\pm$ 0.02}	& 0.20{\scriptsize$\pm$ 0.04}	& 0.18{\scriptsize$\pm$ 0.02}	& -0.13{\scriptsize$\pm$ 0.05}  \\ 
\cline{2-7}
&  \multirow[t]{2}{*}{IIT-DAS}  &  Full	& 0.18{\scriptsize$\pm$ 0.06}	& 0.21{\scriptsize$\pm$ 0.03}	& 0.12{\scriptsize$\pm$ 0.08}	& -0.14{\scriptsize$\pm$ 0.06}  \\ 
&    &  LoRA	& 0.13{\scriptsize$\pm$ 0.05}	& 0.18{\scriptsize$\pm$ 0.04}	& 0.11{\scriptsize$\pm$ 0.05}	& -0.09{\scriptsize$\pm$ 0.06}  \\ 
\midrule
\multirow[t]{4}{*}{\emph{Sighted}} & None & None & 0.14 &  & 0.11 & $-$0.08 \\
\cline{2-7}
\emph{(with image)} & \multirow[t]{2}{*}{Behavioral}  &  Full	& 0.26{\scriptsize$\pm$ 0.05}	&	& 0.22{\scriptsize$\pm$ 0.04}	& 0.03{\scriptsize$\pm$ 0.05}  \\ 
&    &  LoRA	& 0.25{\scriptsize$\pm$ 0.02}	&	& 0.19{\scriptsize$\pm$ 0.02}	& -0.04{\scriptsize$\pm$ 0.04}  \\ 
\cline{2-7}
&  \multirow[t]{2}{*}{IIT-DAS}  &  Full	& 0.25{\scriptsize$\pm$ 0.06}	&	& 0.17{\scriptsize$\pm$ 0.07}	& -0.04{\scriptsize$\pm$ 0.07}  \\ 
&    &  LoRA	& 0.22{\scriptsize$\pm$ 0.06}	&	& 0.15{\scriptsize$\pm$ 0.05}	& 0.02{\scriptsize$\pm$ 0.06}  \\ 
\bottomrule
\end{tabular}
}
\caption{Correlation between model similarity scores and human preferences \cite{kreiss2022context}.}
\label{tab:app-human-correlation}
\end{table}

\paragraph{Results and Discussion}

For each model evaluated, we report correlations averaged across 5 runs. Table \ref{tab:transfer} shows the correlation between BLV individuals' preferences and model similarity scores for the \textit{overall} and \textit{imaginability} dimensions. 

Our results show clearly that fine-tuning \clip\ on the Concadia dataset results in a \clipscore\ that is better aligned with the judgments of BLV individuals. This agrees with the finding that the description--caption distinction is important for BLV users \cite{kreiss2022context}. 

A broad trend is that the more a model is able to distinguish between descriptions and captions the more it aligns with the judgements of BLV individuals. As such, the models trained on the behavioral objective have the highest correlations. 

Table \ref{tab:app-human-correlation} reports the correlation between fine-tuned \clipscores\ and evaluations from BLV individuals, sighted individuals without access to the image the text describes, and sighted individuals with access to the image.

We find that fine-tuning \clip\ on the Concadia dataset improves the model's correlation with human judgements of text descriptions. We note that fine-tuning \clip\ does not significantly improve the model's correlation with human judgements of irrelevant details in the text description. This makes sense, because our training scheme did not take into account to the context of the Concadia image. 

\begin{figure*}[htbp]
    \centering
    \begin{subfigure}[b]{0.33\textwidth}
    \centering
    \resizebox{\textwidth}{!}{
    \begin{tabular}{llcc}
    \toprule
    \textbf{Training} & \textbf{Mediation} & \textbf{Concreteness} & \textbf{Imageability} \\
    \midrule
    \multirow{3}{*}{None} & None & 0.20 & 0.16 \\
    & Through & 0.14 & 0.07 \\
    & Around & - & - \\
    \midrule
    \multirow{3}{*}{Behavioral} & None & $0.26 \pm 0.08$ & $0.31 \pm 0.08$ \\
    & Through & $0.24 \pm 0.08$ & $0.24 \pm 0.08$ \\
    & Around & $0.24 \pm 0.02$ & $0.30 \pm 0.03$ \\
    \midrule
    \multirow{3}{*}{IIT-DAS} & None & $0.26 \pm 0.08$ & $0.23 \pm 0.08$ \\
    & Through & $0.23 \pm 0.03$ & $0.23 \pm 0.03$ \\
    & Around & $0.10 \pm 0.09$ & $0.03 \pm 0.09$ \\
    \bottomrule
    \end{tabular}
    }
    \caption{Correlation between integrated gradient attributions and per-token human labels for concreteness and imageability. The error bounds are 95\% confidence intervals from runs with five random seeds.}
    \label{tab:ig-correlation}
    \end{subfigure}
    \hfill
    \begin{subfigure}[b]{0.64\textwidth}
    \centering
    \begin{minipage}{0.27\textwidth}
    \resizebox{!}{\textwidth}{
    \includegraphics{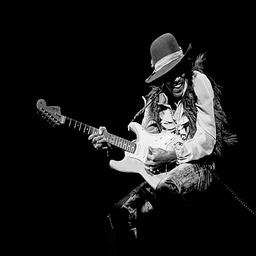}
    }
    \end{minipage}
    \begin{minipage}{0.7\textwidth}
    \resizebox{\textwidth}{!}{
      \centering
      \setlength{\tabcolsep}{4pt}
        \begin{tabular}{@{} lcl @{}}
        \toprule
        \textbf{Example} & \textbf{Mediation} & \textbf{Attribution} \\
        \midrule
        \multirow{5}{*}{Desc.} & \multirow{2}{*}{\xmark} & \colorbox{green!33}{\strut a}\colorbox{green!100}{\strut black}\colorbox{green!46}{\strut and}\colorbox{green!53}{\strut white}\colorbox{green!26}{\strut photograph}\colorbox{green!15}{\strut of}\colorbox{green!36}{\strut jimi}\colorbox{green!73}{\strut hendrix}\\
        & & \colorbox{green!27}{\strut playing}\colorbox{green!5}{\strut a}\colorbox{green!3}{\strut fender}\colorbox{green!26}{\strut strato}\colorbox{green!29}{\strut caster}\colorbox{green!7}{\strut electric}\colorbox{green!44}{\strut guitar} \\
        & \multirow{2}{*}{\cmark} &  \colorbox{green!34}{\strut a}\colorbox{green!100}{\strut black}\colorbox{green!50}{\strut and}\colorbox{green!51}{\strut white}\colorbox{green!39}{\strut photograph}\colorbox{green!12}{\strut of}\colorbox{magenta!6}{\strut jimi}\colorbox{magenta!6}{\strut hendrix}\\
       & &  \colorbox{green!21}{\strut playing}\colorbox{green!7}{\strut a}\colorbox{green!2}{\strut fender}\colorbox{green!10}{\strut strato}\colorbox{green!16}{\strut caster}\colorbox{green!10}{\strut electric}\colorbox{green!41}{\strut guitar} \\
        \midrule
        \multirow{2}{*}{Caption} & \xmark & \colorbox{green!58}{\strut jimi}\colorbox{green!100}{\strut hendrix}\colorbox{magenta!10}{\strut ,}\colorbox{magenta!70}{\strut fillmore}\colorbox{magenta!29}{\strut east}\colorbox{magenta!9}{\strut ,}\colorbox{magenta!9}{\strut may}\colorbox{magenta!17}{\strut 1}\colorbox{magenta!10}{\strut 0}\colorbox{magenta!10}{\strut ,}\colorbox{magenta!1}{\strut 1}\colorbox{magenta!11}{\strut 9}\colorbox{magenta!15}{\strut 6}\colorbox{magenta!5}{\strut 8} \\
        & \cmark & \colorbox{magenta!1}{\strut jimi}\colorbox{magenta!15}{\strut hendrix}\colorbox{magenta!8}{\strut ,}\colorbox{magenta!100}{\strut fillmore}\colorbox{magenta!43}{\strut east}\colorbox{magenta!9}{\strut ,}\colorbox{magenta!12}{\strut may}\colorbox{magenta!20}{\strut 1}\colorbox{magenta!6}{\strut 0}\colorbox{magenta!8}{\strut ,}\colorbox{green!6}{\strut 1}\colorbox{magenta!8}{\strut 9}\colorbox{magenta!18}{\strut 6}\colorbox{magenta!4}{\strut 8} \\
        \midrule
        \end{tabular}
    }
    \end{minipage}
    \begin{minipage}{0.27\textwidth}
    \resizebox{!}{\textwidth}{
    \includegraphics{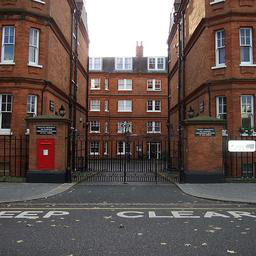}
    }
    \end{minipage}
    \begin{minipage}{0.7\textwidth}
    \resizebox{\textwidth}{!}{
      \centering
      \setlength{\tabcolsep}{4pt}
        \begin{tabular}{lcl}
        \phantom{\textbf{Example}} & \phantom{\textbf{Mediation}} &  \phantom{\textbf{Attribution}} \\
        \midrule
        \multirow{2}{*}{Desc.} & \xmark & \colorbox{green!14}{\strut a}\colorbox{magenta!88}{\strut block}\colorbox{green!53}{\strut of}\colorbox{green!100}{\strut flats}\colorbox{magenta!17}{\strut behind}\colorbox{magenta!36}{\strut a}\colorbox{magenta!32}{\strut set}\colorbox{magenta!6}{\strut of}\colorbox{magenta!8}{\strut high}\colorbox{green!21}{\strut security}\colorbox{green!77}{\strut gates} \\
        & \cmark  & \colorbox{green!6}{\strut a}\colorbox{magenta!100}{\strut block}\colorbox{green!47}{\strut of}\colorbox{green!74}{\strut flats}\colorbox{magenta!9}{\strut behind}\colorbox{magenta!13}{\strut a}\colorbox{magenta!21}{\strut set}\colorbox{magenta!1}{\strut of}\colorbox{magenta!20}{\strut high}\colorbox{magenta!42}{\strut security}\colorbox{green!10}{\strut gates} \\
        \midrule
        \multirow{5}{*}{Caption} & \multirow{2}{*}{\xmark} &  \colorbox{magenta!87}{\strut the}\colorbox{magenta!64}{\strut exclusive}\colorbox{magenta!95}{\strut block}\colorbox{green!11}{\strut of}\colorbox{green!0}{\strut flats}\colorbox{magenta!100}{\strut in}\colorbox{magenta!19}{\strut chelsea}\colorbox{magenta!53}{\strut ,}\colorbox{green!28}{\strut london}\\
        & & \colorbox{magenta!54}{\strut that}\colorbox{magenta!44}{\strut were}\colorbox{magenta!25}{\strut used}\colorbox{magenta!11}{\strut as}\colorbox{magenta!31}{\strut the}\colorbox{magenta!3}{\strut exterior}\colorbox{magenta!7}{\strut of}\colorbox{magenta!56}{\strut mark}\colorbox{magenta!59}{\strut 's}\colorbox{green!64}{\strut flat} \\
        & \multirow{2}{*}{\cmark} &  \colorbox{magenta!75}{\strut the}\colorbox{magenta!59}{\strut exclusive}\colorbox{magenta!100}{\strut block}\colorbox{green!7}{\strut of}\colorbox{magenta!27}{\strut flats}\colorbox{magenta!68}{\strut in}\colorbox{magenta!48}{\strut chelsea}\colorbox{magenta!47}{\strut ,}\colorbox{magenta!35}{\strut london}\\ 
        & & \colorbox{magenta!41}{\strut that}\colorbox{magenta!38}{\strut were}\colorbox{magenta!30}{\strut used}\colorbox{magenta!10}{\strut as}\colorbox{magenta!27}{\strut the}\colorbox{magenta!18}{\strut exterior}\colorbox{magenta!9}{\strut of}\colorbox{magenta!47}{\strut mark}\colorbox{magenta!52}{\strut 's}\colorbox{green!48}{\strut flat} \\
        \bottomrule
        \end{tabular}
    }
    \end{minipage}
    \caption{Integrated gradient attributions for the IIT-DAS model run on an image and its corresponding description and caption from the Concadia dataset. A positive token attribution indicates a positive impact on \clipscore\ (green), and negative token attribution indicates a negative impact (magenta). 
    }
    \label{fig:ig-attribution}
    \end{subfigure}
    \caption{Mediated integrated gradient results.}
    \label{fig:mainfig}
\end{figure*}

\subsection{Integrated Gradients}

A key benefit of localizing the description--caption distinction in \clip\ with IIT-DAS is that we can interpret \clip's representation of a text's communicative purpose (description or caption) separately from \clip's similarity score. In this section, we conduct an analysis of how \clip\ distinguishes between descriptions and captions using an attribution method called integrated gradients (IG; \citealt{sundararajan2017axiomatic}) that evaluates the contribution of each text token to the output \clipscore.

We are particularly curious about how tokens impact the representation of the description--caption distinction. To answer this question, we \emph{mediate} the gradient computation through the linear subspace learned by IIT-DAS \cite{wu2023causal}. We hypothesize that, since the intervention site is trained to represent the underlying purpose of a text (i.e.\ description or caption), gradient attributions that are mediated \textit{through} the linear subspace learned by IIT-DAS will pick out tokens that highlight the description--caption distinction. 

Figure~\ref{fig:ig-attribution} shows an example image from Concadia and the IG attributions of its corresponding description and caption on the IIT-DAS model. We observe that although the overall attributions are positive for the guitarist's name (``Jimi Hendrix''), the \emph{mediated} attributions for these tokens are negative.
While ``Jimi Hendrix'' is aligned with the image (high overall), proper names are less likely to appear in descriptions (low mediated).

% We hypothesize that the interpretability afforded by mediated IG attributions will align with the distinct purposes behind describing and captioning images. Specifically, descriptions are easier to visualize than captions, since their goal is to supplant the image's visual components as opposed to supplement them \cite{kreiss2022concadia}. Hence, we expect that words with higher attributions are easier to visualize.  And, indeed, we find that mediated IG attributions for \clip\ trained with IIT-DAS correlate with human ratings for \textit{imageability} and \textit{concreteness}, whereas IG attributions for the original \clip\ model do not. See Appendix \ref{app:IG} for more details on mediated IG as well as imageability and concreteness correlations.

\paragraph{Dataset} 
We hypothesize that the interpretability afforded by mediated integrated gradients will align with the distinct purposes behind describing and captioning images. Specifically, descriptions are easier to visualize than captions, since their goal is to supplant the image's visual components as opposed to supplement them \cite{kreiss2022concadia}. Hence, we expect that words with higher integrated gradient attributions are easier to visualize. 

We consult two collections of human ratings for visualization-related concepts. The first dataset consists of 5,500 words rated by \emph{imageability}, or how well a word evokes a clear mental image in the reader's mind \cite{scott2019glasgow}. The second dataset consists of over 40,000 words rated by \emph{concreteness}, or how clearly a word corresponds to a perceptible entity \cite{brysbaert2014concreteness}.\footnote{Although imageability and concreteness are slightly different concepts, the imageability and concreteness ratings have a correlation factor of 0.88 with each other.} We randomly sample 100 captions and 100 descriptions from the test split of the Concadia dataset that contain at least one word within both of our datasets, consisting of 420 unique tokens in total. We compute the integrated gradient attributions for all tokens in those sentences, and report their correlations with imageability and concreteness ratings. 

\paragraph{Results and Discussion}

Table~\ref{tab:ig-correlation} displays the correlations between token-level attributions of the LoRA model output and human ratings for imageability and concreteness. All fine-tuning methods achieve a stronger correlation with imageability and concreteness ratings than the base \clip\ model. 

Although all fine-tuning methods result in mediated gradient attributions that correlate with imageability and concreteness, only the IIT-DAS attributions localize to the mediation site. The difference in mediating through vs. around the learned site is significant for the IIT-DAS model ($0.23 \pm 0.03$ vs. $0.10 \pm 0.09$ for concreteness, and $0.23 \pm 0.03$ vs. $0.03 \pm 0.09$ for imageability). 

Our results show that fine-tuning \clip\ to prefer descriptions over captions with IIT-DAS results in models whose attributions correspond to the human-interpretable concept of imageability and concreteness. We also find that mediating integrated gradients through the representation targeted by IIT-DAS preserves this correlation and allows for an analysis of which tokens contribute to distinguishing descriptions from captions.

\section{Conclusion}
We update the \clip\ model to prefer descriptions over captions using Concadia and produce a useful, accessible, and interpretable model.

\section*{Limitations}
Our results serve as proof concept for using IIT-DAS to update \clip\ with the Concadia dataset. This is one model and one dataset, so general conclusions about the use of IIT-DAS for updating a pretrained model should not be drawn. We hope future work will shed further light on the value of IIT-DAS.

The Concadia dataset provides textual context for each image-description-caption triple. We do not use the context in our experiments, but we are excited about future work that incorporates this data. Whereas our work focuses on the specific purposes of describing and captioning an image, the context of an image can illuminate many other purposes (e.g.\ search, geolocation, social communication) and models that incorporate it can enrich our work. 

\section*{Ethics Statement}
We believe that modern AI is a transformative technology that should benefit all of us and accessibility applications are an important part of this.

\bibliography{anthology,custom}
\bibliographystyle{acl_natbib}

\appendix

\newpage

\section{Causal Models, Interchange Intervention Training, and Distributed Alignment Search}\label{app:iit}
This section follows \citealt{geiger2023causal,geiger2023finding}.

\textbf{Causal Models} can represent a variety of processes, including deep learning models and symbolic algorithms. A causal model $\mathcal{M}$ consist of variables $\mathbf{V}$, and, for each variable $X \in \mathbf{V}$, a set of values $\texttt{Val}(X)$, and a structural equation $F_X:\texttt{Val}(\mathbf{V}) \to \texttt{Val}(X)$, which is a function that takes in a setting of all the variables and outputs a value for $X$. The solutions of a model $\mathcal{M} = (\mathbf{V}, \texttt{Val}, F)$ are settings for all variables $\mathbf{v} \in \texttt{Val}(\mathbf{V})$ such that the output of the causal mechanism $F_X(\mathbf{v})$ is the same value that $\mathbf{v}$ assigns to $X$, for each $X \in \mathbf{V}$.

We only consider structural causal models with a single solution that induces a directed acyclic graphical structure such that the value for a variable $X$ depends only on the set of variables that point to it, denoted as its \emph{parents} $\emph{PA}_X$. Because of this, we treat each causal mechanism $F_X$ as a function from parent values in $\texttt{Val}(\emph{PA}_X)$ to a value in $\texttt{Val}(X)$. We denote the set of variables with no parents as $\mathbf{V}_{\emph{in}}$ and those with no children $\mathbf{V}_{\emph{out}}$. 

Given $\mathsf{input} \in \texttt{Val}(\mathbf{V}_{\emph{in}}$) and variables $\mathbf{X} \subseteq \mathbf{V}$, we define $\texttt{GET}(\mathcal{M}, \mathsf{input}, \mathbf{X}) \in \texttt{Val}(\mathbf{X})$ to be the setting of $\mathbf{X}$ determined by the given $\mathsf{input}$ and model $\mathcal{M}$. For example, $\mathbf{X}$ could correspond to a hidden activation layer in a neural network, and $\texttt{GET}(\mathcal{M}, \mathsf{input}, \mathbf{X})$ then denotes the particular values that $\mathbf{X}$ takes on when the model $\mathcal{M}$ processes $\mathsf{input}$.

\textbf{Interventions} simulate counterfactual states in causal models. For a set of variables $\mathbf{X}$ and a setting for those variables $\mathbf{x} \in \texttt{VAL}(\mathbf{X})$, we define $\mathcal{M}_{\mathbf{X} \gets \mathbf{x}}$ to be the causal model identical to $\mathcal{M}$, except that the structural equations for $\mathbf{X}$ are set to constant values $\mathbf{x}$. In the case of neural networks, we overwrite the activations with $\mathbf{x}$ in-place so that gradients can back-propagate through $\mathbf{x}$. 

\textbf{Distributed interventions} also simulate counterfactual states in causal models, but do so by editing the causal mechanisms rather than overwriting them to be a constant. Given variables $\mathbf{X}$ and an invertible function $\rmat:\texttt{VAL}(\mathbf{X}) \to \texttt{VAL}(\mathbf{Y})$ mapping $\mathbf{X}$ into a new variable space $\mathbf{Y}$, define $\rho(\mathcal{M})$ to be the model where the variables $\mathbf{X}$ are replaced with the variables $\mathbf{Y}$. For a setting of the new variable space $\mathbf{y} \in \texttt{VAL}(\mathbf{Y})$, it follows that $\rmat^{-1}(\rmat(\mathcal{M})_{\mathbf{Y} \gets \mathbf{y}})$ is the causal model identical to $\mathcal{M}$, except that the causal mechanisms for $\mathbf{X}$ are edited to fix the value of $\mathbf{Y}$ to $\mathbf{y}$. If $\rmat$ is differentiable, then gradients back-propagate through $\mathbf{y}$. 

A \textbf{distributed interchange intervention} fixes variables to the values they would have taken if a different input were provided. Consider a causal model $\mathcal{M}$, an invertible function $\rmat:\texttt{VAL}(\mathbf{X}) \to \texttt{VAL}(\mathbf{Y})$, source and base inputs $\mathbf{s},\mathbf{b} \in \texttt{Val}(\mathbf{V}_{\emph{in}})$, and a set of intermediate variables $\mathbf{X} \subset \mathbf{V}$. A distributed interchange intervention computes the value $\mathbf{V}_{\emph{out}}$ when run on $\mathbf{b}$, intervening on the (distributed) intermediate variables $\mathbf{Y}$ to be the value they take on when run on $\mathbf{s}$. Formally, we define 
\begin{multline*}
\dintinv(\mathcal{M}, \rmat, \mathbf{b}, \mathbf{s}, \mathbf{Y}) =\\ 
\texttt{GET}(\rmat^{-1}(\rmat(\mathcal{M})_{ \mathbf{Y} \gets \texttt{GET}(\rmat(\mathcal{M}), \mathbf{s}, \mathbf{Y})}), \mathbf{b}, \mathbf{V}_{\emph{out}})
\end{multline*}

\paragraph{High-Level Models} Define a class $\Delta$ to contain only causal models that consist of the following three variables. The input variable $X$ takes on the value of some image-text pair in Concadia $(x_{\text{image}}, x_{\text{text}})$; the intermediate variable $P$ (i.e.\ purpose) takes on a value from $\{ \text{``describe''}, \text{``caption''}\}$ depending on the Concadia label for $X$; and the output variable $Y$ takes on a real value that represents the similarity between $x_{\text{image}}$ and $x_{\text{text}}$. The causal mechanism of $Y$ must be such that for every image, the description text is assigned a higher \clipscore\ than the caption text. If a \clip\ model implements any algorithm in $\Delta$, then it will assign descriptions higher scores than captions.

\section{Training Details}
\label{app:training-details}
In this paper, we propose a novel combination of LoRA fine-tuning with the IIT-DAS objective. We apply DAS to the representation after LoRA is applied (i.e., $Wx + W'x$, where $W$ is the original matrix weight and $W'$ is the low-rank adaptation), and fine-tune the LoRA parameters $W'$ and the rotation matrix $\rho^{\theta}$.

We fine-tune the \clip\ \texttt{ViT-B/32} Transformer model released by OpenAI\footnote{\url{https://huggingface.co/openai/clip-vit-base-patch32}}, which consists of 12 transformer layers with a hidden dimension of 512, constituting $\sim$150M overall parameters. For all fine-tuning runs, we use Adam optimization with default parameters \cite{kingma2014adam} and a batch size of 12.

\paragraph{Behavioral Objective} 
We fine-tune \clip\ on the training split of the Concadia dataset (77,534 datapoints) with early-stopping validation on the Concadia validation split (9,693 datapoints). We conduct a hyperparameter grid search over a learning rate $\emph{lr} \in \{10^{-3}, 0.5\cdot10^{-3}, 10^{-4}, 0.5\cdot10^{-4}, \ldots, 10^{-6}, 0.5\cdot10^{-6}\}$, an L2 normalization factor $\emph{l}_2 \in \{0, 0.1, 0.01, 0.001 \}$. We select the configuration with the highest accuracy on the Concadia validation split within 5 epochs ($\emph{lr} = 0.5\cdot10^{-6}$, $\emph{l}_2 = 0$). A training run takes around 3 hours on an RTX A6000 NVIDIA GPU. 

\paragraph{IIT-DAS Objective}
We fine-tune \clip\ on 100,000 triplets sampled from the train split of the Concadia dataset (out of $77,534 \times 77,534$ possible caption--description pairs). We conduct a hyperparameter grid search over a learning rate $\emph{lr} \in \{10^{-5}, 5^{-6}, 10^{-6}\}$, as well as over the intervention site: the layer $\emph{layer} \in \{6, 8, 10\}$, the intervention site size $\emph{intervention-size} \in \{32, 64, 128, 256\}$. We select the configuration with the highest accuracy on the Concadia validation split within 5 epochs ($\emph{lr} = 10^{-5}, \emph{layer} = 10, \emph{intervention-size} = 256$). A training run takes around 6 hours on an RTX A6000 NVIDIA GPU. 

\paragraph{LoRA Fine-Tuning}
We perform an additional hyperparameter search for low-rank fine-tuning of \clip\ for both the behavioral and IIT-DAS objective. We take the best configuration for full-finetuning, and then perform a search over the LoRA rank $\emph{rank} \in \{8, 16, 32, 64, 128\}$, the LoRA dropout $\emph{dropout} \in \{0, 0.1, 0.01\}$, and whether to apply LoRA to all linear layers, all attention weights, or only the query and value projection matrices within attention weights. We select the configuration with the highest accuracy on the Concadia validation split with 5 epochs. For the behavioral objective, the configuration is $\emph{rank} = 64$, $\emph{dropout} = 0$, with LoRA applied to all attention weights. For the IIT-DAS objective, the configuration is the same but with $\emph{rank} = 128$.

\paragraph{Joint Objective}
We note that the behavioral objective and IIT-DAS objective can complement each other -- the former teaches the model to prefer descriptions to captions, and the latter teaches the model to localize this distinction in a particular representation. Hence, we consider a joint objective, where we train the model to minimize $$\mathcal{L}_\text{Joint} = \alpha \mathcal{L}_\text{IIT} + (1 - \alpha)\mathcal{L}_\text{Behavioral}$$ We search over an interpolation factor $\alpha \in \{0.2, 0.3, 0.5, 0.7, 0.8\}$. However, we find that the joint objective neither improves upon the behavioral objective nor the IIT-DAS objective. 

\section{Transfer Evaluations}
\label{app:transfer-tasks}

\begin{figure}
    \centering
    \includegraphics[width=\linewidth]{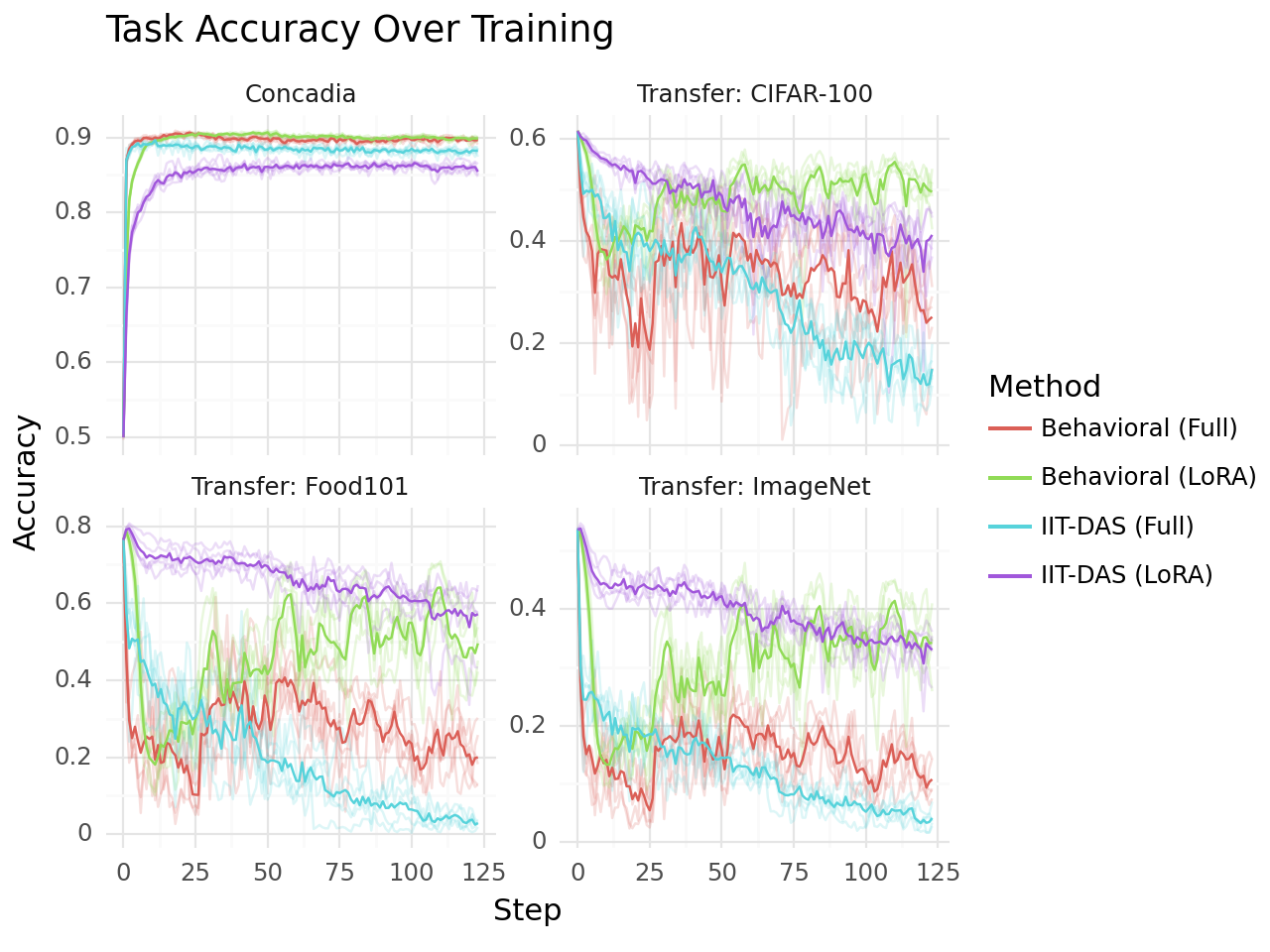}
    \caption{Accuracy on the Concadia test set, and the three transfer tasks selected for transfer evaluation (CIFAR-100, Food101, and ImageNet).}
    \label{fig:accuracy-over-training}
\end{figure}

We evaluate our fine-tuned \clip\ models on tasks selected from the original suite of zero-shot evaluations performed on \clip\ \cite{Radford}. Specifically, we choose three zero-shot image classification tasks in which \clip\ has strong performance, described briefly below.

\paragraph{CIFAR-100}
Images from 100 different categories \cite{krizhevsky2009learning}.

\paragraph{Food101}
Food images labeled from 101 categories \cite{bossard14}. 

\paragraph{ImageNet} 
Image-text pairs for each synonym set in the WordNet hierarchy \cite{imagenet15russakovsky, miller1995wordnet}.

Pre-trained \clip\ varies greatly in its ability to generalize to each of these tasks, but it does outperform a supervised linear classifier trained on ResNet-50 features \cite{Radford}. We report the macro-averaged F1 score on zero-shot classification for each of the transfer tasks listed above, averaged across 5 randomly seeded training runs. During evaluation, we prefix ``An image of \_\_\_'' to each label in order to improve zero-shot generalization.

Figure \ref{fig:accuracy-over-training} shows model accuracy throughout training for the 5 randomly seeded training runs of each training objective and fine-tuning method. Although the training converges quickly on the Concadia dataset, training for longer seems to allow the model to recover performance on the evaluated transfer tasks when using LoRA fine-tuning. Hence, for the evaluation scores reported in Table \ref{tab:transfer}, we train models well past convergence for 10 epochs, and then select a model to balance the trade-off between Concadia accuracy and transfer capabilities:
\begin{itemize}
    \item Compute a recovery percentage: divide the model's accuracy on the transfer task by the accuracy of pre-trained \clip\ on that task (see Table \ref{tab:transfer}).
    \item Compute a transfer score: average the recovery percentage across the three transfer tasks.
    \item Compute accuracy--transfer trade-off score: compute $\alpha \cdot$ (Concadia accuracy) $+ (1 - \alpha) \cdot$ (transfer score).
    \item For each seeded run, select the training step with the highest accuracy--transfer trade-off score.
\end{itemize}

\begin{figure}
    \centering
    \includegraphics[width=\linewidth]{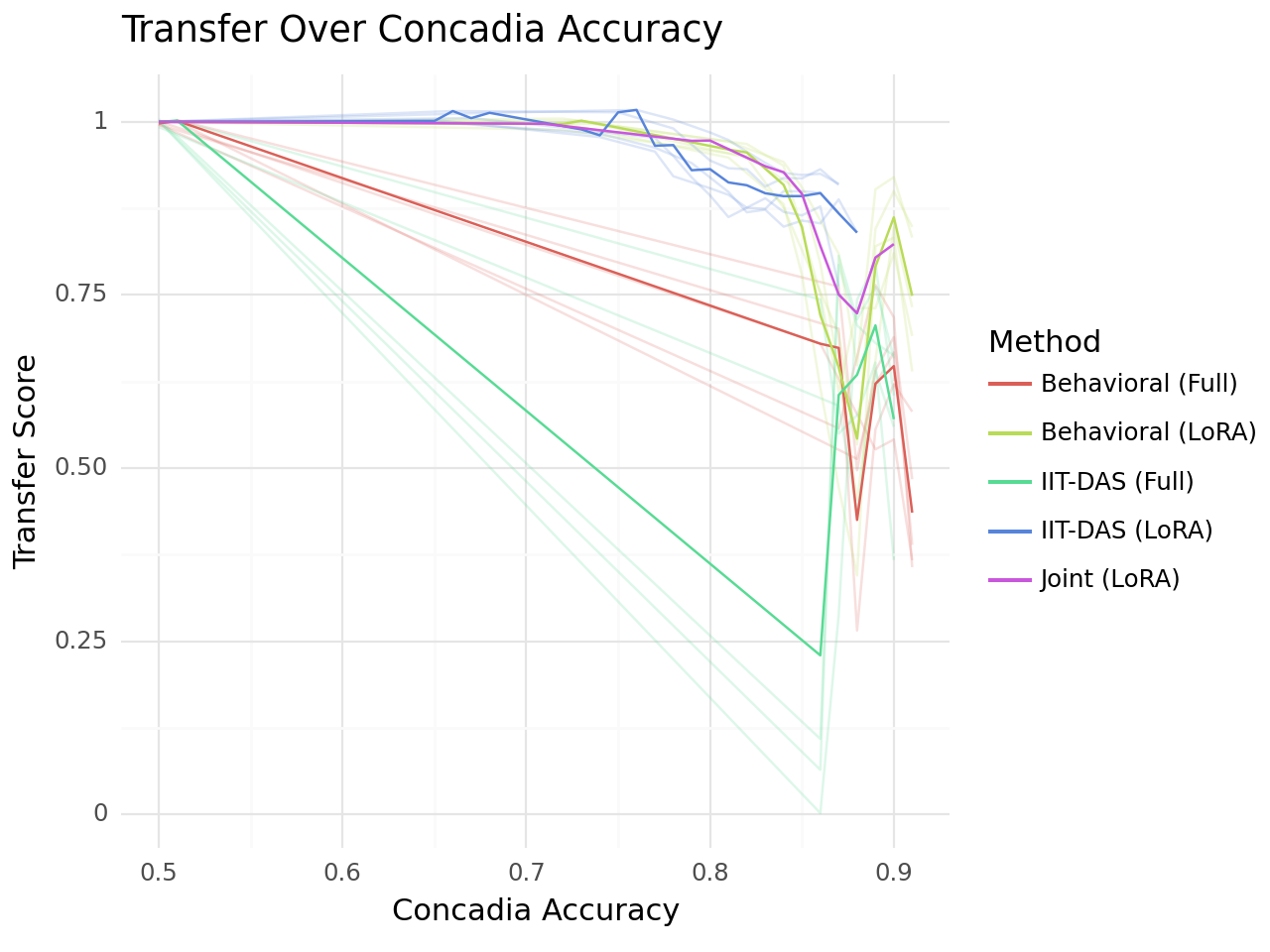}
    \caption{Transfer score (averaged recovery percentage over all transfer tasks) over accuracy on the Concadia test set. A point on a seeded run yields a trade-off between sensitivity to the caption--description distinction and preserving the capabilities of \clip. The joint objective refers to a training run minimizing both the behavioral and the IIT-DAS objective (see Appendix~\ref{app:training-details}).}
    \label{fig:transfer-over-accuracy}
\end{figure}

We manually pick a trade-off of $\alpha = 0.9$, meaning we weigh the trade-off as 90\% Concadia accuracy and 10\% transfer accuracy; we find that trade-offs with a lower $\alpha$ result in poor Concadia accuracy. 

We visualize other possible trade-off in Figure \ref{fig:transfer-over-accuracy}. We find that the strongest trade-off for IIT-DAS with LoRA fine-tuning is $\approx86\%$ accuracy on the Concadia test set and $\approx82\%$ transfer score. Meanwhile, the strongest trade-off for the behavioral objective with LoRA fine-tuning is skewed towards stronger Concadia performance, with $\approx90\%$ accuracy on Concadia and $\approx80\%$ transfer score.

Figure \ref{fig:transfer-over-accuracy} also displays the transfer--accuracy trade-off for a joint objective that minimizes both the behavioral and the IIT-DAS objective (see Appendix~\ref{app:training-details} for details). The joint objective seems to strike some balance between the trade-off curves of the behavioral and IIT-DAS objectives -- maintaining higher transfer scores around $\approx85\%$ Concadia accuracy compared to the behavioral objective, and reaching higher Concadia accuracy ($\approx90\%$) compared to IIT-DAS. Nevertheless, for any given Concadia accuracy value, one of the behavioral or IIT-DAS objectives achieves at least as high a transfer score as the joint objective. We leave strategies to optimally combine the behavioral and IIT-DAS objectives for further study.

\section{BLV and Sighted Evaluations}
\label{app:human-correlations}

\citet{kreiss2022context} recruited 16 participants via
email lists for BLV users that were unaware of the studies purpose. Participants were totally blind (7), nearly blind (3), light perception only (5), and low vision (1). 15 participants reported relying on screen readers (almost) always when browsing
the Web, and one reported using them often. In total, judgements were provided for 68 descriptions, comprising 18 images and 17 Wikipedia articles. 

\section{Integrated Gradients}\label{app:IG}
Given a model $\mathcal{C}^{\theta}$ with input $x$ and baseline $x'$, the integrated gradient attributions of $x$ along its $i$th dimension are computed as follows. 
\begin{multline}
    \texttt{IntegratedGrads}_i(x) = (x_i - x'_i) \cdot \\
    \int_{\alpha=0}^1 \frac{\partial \mathcal{C}^{\theta}(x' + \alpha (x - x'))}{\partial x_i} \ \partial \alpha.
\end{multline}
\paragraph{Mediated Integrated Gradients}
Let $H$ be the activation of $\mathcal{C}^{\theta}$ at the intervention site when run on input $x$. Our mediated integrated gradient is \[
\frac{\partial \mathcal{C}^{\theta}}{\partial x_i} \text{ mediated by } H = \frac{\partial \mathcal{C}^{\theta}}{\partial H} \times \frac{\partial H}{\partial x_i}.
\] Unlike $\frac{\partial \mathcal{C}^{\theta}}{\partial x_i}$, which computes the gradient of the model output with respect to the input $x_i$, the mediated gradient only flows through the intervention site $H$. We compute mediated integrated gradients by applying this gradient method within the integral of the integrated gradients equation.

\end{document}